\documentclass[11pt,a4paper]{article}
\usepackage[hyperref]{naaclhlt2019}
\usepackage{times}
\usepackage{latexsym}
\usepackage{url}

\usepackage{amsmath}
\usepackage{amsthm}
\usepackage{amssymb}
\usepackage{cases}
\usepackage{algorithm}
\usepackage{algorithmic}
\usepackage{multirow}
\usepackage{graphicx}
\usepackage[caption=false]{subfig}
\usepackage{footnote}
\usepackage{mdwlist}
\usepackage{enumerate}
\usepackage{threeparttable}

\graphicspath{{./}}
\DeclareGraphicsExtensions{.png,.pdf}

\aclfinalcopy 

\title{Forward and Backward Knowledge Transfer for Sentiment Classification}

\author{Hao Wang$^{\S,\P}$, Bing Liu$^{\P}$, Shuai Wang$^{\P}$, Nianzu Ma$^{\P}$, Yan Yang$^{\S}$ \\
  $^{\S}$School of Information Science and Technology, Southwest Jiaotong University \\
  {\tt cshaowang@gmail.com, yyang@swjtu.edu.cn} \\
  $^{\P}$Department of Computer Science, University of Illinois at Chicago \\
  {\tt liub@uic.edu, gshuaishuai@gmail.com, jingyima005@gmail.com} \\}

\date{}
\begin{document}
\maketitle
\begin{abstract}
This paper studies the problem of learning a sequence of sentiment classification tasks. The learned knowledge from each task is retained and used to help future or subsequent task learning. This learning paradigm is called \text{lifelong learning} (LL). However, existing LL methods either only transfer knowledge forward to help future learning and do not go back to improve the model of a previous task or require the training data of the previous task to retrain its model to exploit backward/reverse knowledge transfer. This paper studies reverse knowledge transfer of LL in the context of na\"ive Bayesian (NB) classification. It aims to improve the model of a previous task by leveraging future knowledge without retraining using its training data. This is done by exploiting a key characteristic of the generative model of NB. That is, it is possible to improve the NB classifier for a task by improving its model parameters directly by using the retained knowledge from other tasks. Experimental results show that the proposed method markedly outperforms existing LL baselines.

\end{abstract}

\section{Introduction}
Lifelong learning (LL) aims to learn a sequence (possibly never ending) of tasks. After a task is learned, its knowledge is retained and later used to help future task learning \cite{thrun1998lifelong,silver2013lifelong,chen2018lifelong}. This paper studies LL for sentiment classification (SC). SC classifies an opinion document (e.g., a product review) as expressing a positive or negative sentiment \cite{pang2008opinion,cambria2017sentiment}. In the LL setting, we are interested in learning a sequence of SC tasks. However, exiting LL methods mainly help future task learning by leveraging the knowledge learned from past tasks, which we call \textit{forward knowledge transfer} (FKT). If the model of a past task needs to be improved, LL can also use the knowledge learned in future tasks to help, but the past task training data is needed for retraining. 

In this paper, we not only want to achieve FKT to improve any future task learning, but also want to achieve \textit{reverse knowledge transfer} (RKT) to improve the model of any past task without retraining using its training data. This enhanced LL setting is natural as we humans seem to learn in a similar way. We can use some newly learned knowledge to help improve a previous task without re-learning from the past experiences or data, which are often forgotten. 

\textbf{Problem Definition:} At any point in time, a learner has learned $n$ tasks. Each task $T_i$ ($1\!\le\!i\!\le\!n$) has its training data $D_i$. The learned knowledge of each task is retained in a knowledge base (KB). When faced with a new task $T_{n+1}$, the knowledge in the KB is leveraged to help learn $T_{n+1}$. After $T_{n+1}$ is learned, its knowledge is also incorporated into the KB. The knowledge in the KB can also be used to improve the model of any past task $T_i$ with no retraining using its training data $D_i$.




In this paper, we propose to deal with the above problem using na\"ive Bayes (NB) by exploiting its \textit{generative model} parameters. The proposed method is called \textit{L}ifelong NB (LNB). The key idea is that the prior knowledge 
can be mined from the generative model parameters of previous tasks and used to directly revise the generative model parameters of the \textit{target task}, which can be a new or a past task. Since 
the generative model parameters are computed during training for each task and retained, then no retraining will be needed when going back to improve the model of any past task.


In summary, this paper makes the following contributions. (1) It studies an enhanced LL setting, i.e., not only improving future task learning, but also going back to improve the past task models with no retaining using their training data.
(2) It proposes a novel LL model called Lifelong Na\"ive Bayes (LNB) for SC in this new setting by exploiting the parameters of the generative model of NB. To the best of our knowledge, this is the first such formulation.
(3) It evaluates the effectiveness of the proposed model, which shows that the proposed model makes considerable improvement over state-of-the-art baselines.

\section{Related Work}
Most related work to ours is lifelong learning (LL) \cite{mitchell2015never,ruvolo2013ella,chen2015lifelong,fei2016learning,isele2016using,xia2017distantly,ShuXuLiu2017,xu2018lifelong,sun2018active}. However, these works cannot improve the model of a past task using the knowledge learned from other tasks without retraining using the training data of the past task. \newcite{chen2015lifelong} proposed the first LL method (called LSC) for sentiment classification (SC) based on optimization considering the past knowledge. However, it also needs retraining using the past training data to improve its model. \citet{xia2017distantly} presented two LL methods based on voting of individual task classifiers for SC. The first method votes with equal weight for each task classifier, which can be applied to help past tasks. However, this method performs poorly. We will compare with this method experimentally. The second method uses weighted voting, which needs the past task training data for retraining.

Our work is also closely related to 
transfer learning (TL) \cite{Pan2010TLSurvey,Blitzer2007,Andreevskaia2008,Bollegala2011,He2011,Li2012,Li2013,xia2011pos,pan2010cross,Wu2009graph,wu2017active,augenstein2018multi}. However, TL can only uses the source domains to help the target domain, but not the reverse as the target domain has little or no labeled training data. 

\section{Lifelong Na\"ive Bayes}
\subsection{Na\"ive Bayes for Text Classification}
Na\"ive Bayes (NB) text classifier is a generative model \cite{nigam1998learning}. Given a
document $d$ with words $\{w_1,...,w_n\}$, the NB classifier based on multinomial distributions is defined as:
\begin{equation}\label{eq:1}
\footnotesize{
P(c_j|d) = \frac{P(c_j)\prod_{k=1}^{n}P(w_k|c_j)}{\sum_{r=1}^{|C|}P(c_r)\prod_{k=1}^{n}P(w_k|c_r)}
}
\end{equation}
where $c_j$ is positive (+) or negative (-) class in our case, $|C|$ is the number of classes, and the multinomial distribution parameter $P(w_k|c_j)$ is
\begin{equation}
\label{wordPro}
P(w_k|c_j) = \frac{\lambda+N_{c_j,w_k}}{\lambda|V|+\sum_{v=1}^{|V|}N_{c_j,w_v}}
\end{equation}
where $N_{c_j,w_k}$ is the number of times that word $w_k$ occurs in the training documents of class $c_j$, $|V|$ is the vocabulary $V$ size, and $\lambda$ is the smoothing parameter. We use $\lambda\!=\!0.1$ as it is shown in \cite{agrawal2000athena} that $\lambda\!=\!0.1$ (Lidstone smoothing) is superior to $\lambda\!=\!1$ (Laplacian smoothing).

From Eq.~\eqref{wordPro}, we see that NB mainly depends on the word frequency count $N_{c_j,w_k}$, which is also the core \textit{knowledge} that our LNB will retain for each domain in addition to the class prior $P(c_j)$.

\subsection{LNB for Sentiment Classification}




In our LNB, there are three key components: Knowledge Miner (KM), Knowledge Base (KB), and Knowledge-Based Learner (KBL). In brief, KM mines knowledge from training data of each task/domain, KB stores the mined knowledge, and KBL abstracts some high-level knowledge from the KB and leverages it in learning the \textit{target task/domain} $t$, which can be a new domain task or a previous domain task. Following \cite{chen2015lifelong}, we treat the classification in each domain (i.e., a type of product) as a learning task. Thus, we use the terms \underline{\textit{domain}} and \underline{\textit{task}} interchangeably throughout the paper.

The key idea of LNB is to revise the multinomial distribution parameters (i.e., Eq.~\eqref{wordPro}) for the target task using prior
knowledge in KB from the previous tasks to built
a better target task classifier. Below we describe how to make this idea work.

Given a task sequence $\{T_1, T_2, ...\}$, KM extracts two types of knowledge from 
the training data of each new task $T_i$.
(1) $N_{+,w}^{i}$ and $N_{-,w}^{i}$: number of times word $w$ occurs in the training documents of the positive ($+$) and negative ($-$) class in $T_i$, respectively.
(2) $N_{+}^{i}$ and $N_{-}^{i}$: number of training documents in the $+$ and $-$ class in $T_i$, respectively. 
The above two-types of knowledge will be stored in the KB after each new task being learned.

Next, KBL uses the stored knowledge and the newly extracted knowledge by KM (if target task is a new task) to compute the generative model parameters and to build a NB classifier for the target domain $t$. Specially, KBL first abstracts three types of high-level knowledge from the KB:
\vspace{-0.1cm}
\begin{enumerate*}
	\item[(a)] Word-level knowledge $N_{+,w}^{KB}$ and $N_{-,w}^{KB}$: number of times word $w$ occurs in the training documents of the positive (+) and and negative (-) class in all domains except the target domain, i.e., $\!N_{+,w}^{KB}\! =\!\sum\nolimits_{f} {N_{+,w}^{f}}\!$ and $\!N_{-,w}^{KB}\! =\!\sum\nolimits_{f} {N_{-,w}^{f}}$.
	\item[(b)] Target domain-dependent knowledge $Q_{+,w}^{t}$ and $Q_{-,w}^{t}$: ratio of word probability in positive (and negative) class vs. negative (positive) class in the target domain $t$, i.e., $Q_{+,w}^{t}\!=\!\frac{P(w|+)}{P(w|-)}$ and $Q_{-,w}^{t}\!=\!\frac{P(w|-)}{P(w|+)}$. 
	\item[(c)] Domain-level knowledge $M_{+,w}^{KB}$ and $M_{-,w}^{KB}$: number of non-target domains in which $\frac{P(w|+)}{P(w|-)}\!\ge\!\gamma$ and $\frac{P(w|-)}{P(w|+)}\!\ge\!\gamma$, where $P(w|+)$ and $P(w|-)$ are estimated by using Eq.~\eqref{wordPro} in that domain, and $\gamma$ is a parameter. 
	\vspace{-0.1cm}
\end{enumerate*}

Then, KBL integrates these pieces of knowledge to revise $N_{+,w}^t$ and $N_{-,w}^t$ mined from the target domain. We denote the revised results as $\hat{N}_{+,w}^t$ and $\hat{N}_{-,w}^t$. The intuition here is that if a word $w$ can distinguish classes very well in the target domain, we should rely on the target domain. So, we define a set of target domain-dependent words, denoted by $V^t$. A word $w$ belongs to $V^t$ if $Q_{+,w}^{t}\!\ge\!\sigma$ or $Q_{-,w}^{t}\!\ge\!\sigma$, where $\sigma$ is a parameter. 
On the other hand, if a word $w$ is reliable among most non-target domains (e.g., half of the non-target domains), we should follow the knowledge associated with this word in the KB. Similarly, we define a set of domain-reliable words, denoted by $V^{KB}$. A word $w$ belongs to $V^{KB}$ if $M_{+,w}^{KB}\!>\!n/2$ or $M_{-,w}^{KB}\!>\!n/2$, where $n$ is the number of the non-target domains. 



\textbf{Improving past and future domain classification}: It is clear that LNB can treat any past or future domain as the target domain $t$ and improve its classification. LNB only needs the frequency count of each word in each class of each (past or future) domain (which is stored in the KB). Thus, for a past/previous domain, no retraining using its original training data is needed. In summary, our LNB model works for a test document $d_u$ in the target domain as shown in Figure \ref{Alg}.
\begin{figure}[!htb]
	\centering
	\begin{algorithmic}[1]
		\STATE Extract uni-gram features for words  $\{w_1,w_2...,w_n\}$;
		\FOR{each feature word $w_k$}
		\IF{$w_k$ belongs to $V^{KB}$}
		\STATE $\hat{N}_{+,w_k}^t\!=\!R_{w_k}\!\times\! N_{+,w_k}^{KB}$,
		\STATE $\hat{N}_{-,w_k}^t = (1-R_{w_k})\times N_{-,w_k}^{KB}$;\\
		\hspace{-0.3cm}//{~where~$R_{w_k}\!=\!M_{+,w_k}^{KB}/(M_{+,w_k}^{KB}+M_{-,w_k}^{KB})$}
		\ELSIF{$w_k$ belongs to $V^t$}
		\STATE $\hat{N}_{c_j,w_k}^t = N_{c_j,w_k}^t$;
		\ELSE
		\STATE $\hat{N}_{c_j,w_k}^t = N_{c_j,w_k}^{KB} + N_{c_j,w_k}^t$;
		\ENDIF
		\ENDFOR
		\RETURN $argmax_j\ P(c_j|d_u)$.
	\end{algorithmic}
	\vspace{-0.15cm}
	\caption{LNB classification algorithm for a test document $d_u$ in the target domain $t$.}
	\label{Alg}
	\vspace{-0.2cm}
\end{figure}

\section{Experiments}
\textbf{Datasets:} Since the main baseline is the LSC system, we experiment using the same 20 domains dataset\footnote{\url{https://www.cs.uic.edu/~zchen/downloads/ACL2015-Chen-Datasets.zip}} as in~\cite{chen2015lifelong}.~Following \cite{chen2015lifelong}, we use two versions of the dataset with positive and negative classes in different class distributions, i.e., \textit{Natural Class Distribution} and \textit{Balanced Class Distribution}, see supplementary note in Appendix or the work of \citet{chen2015lifelong}. 

We extracted uni-gram features with no feature selection from the raw reviews. Also, we followed \newcite{Pang2002b} to deal with negation words as \newcite{chen2015lifelong}.
\begin{table*}[!htb]
\scriptsize 
\renewcommand\arraystretch{1.2}
\setlength{\tabcolsep}{3.5pt}
	\centering
	\resizebox{1\textwidth}{!}{
	\begin{tabular}{|ccccccccc|ccccccccc|}
		\hline
		\multicolumn{9}{|c|}{\textbf{Average F1-score of the negative classes in the Natural Class Distribution}} & \multicolumn{9}{c|}{\textbf{Average Accuracy of the two classes in the Balanced Class Distribution}} \\
		\hline
		NB-T & NB-S & NB-ST & SVM-T & SVM-S & SVM-ST & LSC & LLV & LNB & NB-T & NB-S & NB-ST & SVM-T & SVM-S & SVM-ST & LSC & LLV & LNB \\
		\hline
		45.20 & 55.00 & 56.49 & 50.39 & 52.66 & 59.15 & 56.62 & 47.46
		& \textbf{64.96} & 77.40 & 74.82 & 80.04 & 76.09 & 75.79 & 79.29 & 82.09 & 78.59
		& \textbf{83.17}
		\\
		\hline
	\end{tabular}}
	\vspace{-0.2cm}
    \caption{New task evaluation: Average sentiment classification (SC) performance over 20 domains.}
	\label{tab:newTask}
\end{table*}
\begin{table*}[!htb]
\scriptsize 
\renewcommand\arraystretch{1.1}
\setlength{\tabcolsep}{3.0pt}
	\centering
	\resizebox{1.0\textwidth}{!}{
\begin{threeparttable}
	\begin{tabular}{|l|ccccccccc|ccccccccc|}
		\hline
		\multirow{2}{*}{\textbf{Sequence}} & \multicolumn{9}{c|}{\textbf{Average F1-score of the negative classes in the Natural Class Distribution}} & \multicolumn{9}{c|}{\textbf{Average Accuracy of the two classes in the Balanced Class Distribution}} \\
		\cline{2-19}
		& NB-T & NB-S & NB-ST & SVM-T & SVM-S & SVM-ST & LSC & LLV & LNB & NB-T & NB-S & NB-ST & SVM-T & SVM-S & SVM-ST & LSC & LLV & LNB \\
		\hline
		S1 &  43.04 & 49.66 & 54.25 & 51.16 & 50.21 & 57.34 & 54.89 & 44.25 & \textbf{63.80} & 78.81 & 71.84 & 77.76 & 76.57 & 73.42 & 80.26 & 81.84 & 79.47 & \textbf{85.26} \\
		S2 &  44.42 & 51.16 & 51.98 & 51.80 & 51.13 & 58.62 & 53.93 & 44.32 & \textbf{63.56} & 78.15 & 71.71 & 78.55 & 76.18 & 69.73 & 78.68 & 81.97 & 80.26 & \textbf{84.74} \\
		S3 &  42.35 & 49.37 & 52.29 & 49.99 & 46.70 & 59.60 & 51.80 & 44.76 & \textbf{63.62} & 78.55 & 73.02 & 79.86 & 75.92 & 71.31 & 78.55 & 81.18 & 79.21 & \textbf{84.60} \\
		S4 &  42.64 & 42.22 & 45.07 & 50.60 & 45.22 & 54.90 & 50.15 & 43.87 & \textbf{63.66} & 78.29 & 70.00 & 78.55 & 75.79 & 69.34 & 76.97 & 79.99 & 79.07 & \textbf{84.74} \\
		S5 &  42.65 & 46.16 & 51.90 & 50.05 & 49.54 & 59.35 & 48.43 & 44.46 & \textbf{63.39} & 78.16 & 71.97 & 77.50 & 75.65 & 71.97 & 80.26 & 80.65 & 78.94 & \textbf{84.87} \\
		S6 &  42.48 & 45.56 & 52.10 & 50.50 & 49.30 & 60.16 & 53.45 & 43.83 & \textbf{63.64} & 78.68 & 68.55 & 80.26 & 75.92 & 67.89 & 76.71 & 81.84 & 79.60 & \textbf{85.26} \\
		S7 &  45.63 & 46.28 & 52.21 & 51.44 & 50.15 & 59.29 & 52.02 & 46.78 & \textbf{65.09} & 78.68 & 69.73 & 77.76 & 75.52 & 69.34 & 77.23 & 79.21 & 79.34 & \textbf{84.47} \\
		S8 &  43.05 & 50.74 & 51.34 & 51.16 & 48.96 & 57.52 & 52.64 & 44.25 & \textbf{63.80} & 78.81 & 71.44 & 80.39 & 76.57 & 71.84 & 79.21 & 82.36 & 79.47 & \textbf{85.26} \\
		S9 &  43.15 & 49.61 & 50.78 & 50.72 & 48.41 & 58.08 & 54.14 & 43.15 & \textbf{63.30} & 78.95 & 71.97 & 79.21 & 76.71 & 71.05 & 80.26 & 81.57 & 79.60 & \textbf{85.26} \\
		S10 &  42.48 & 51.23 & 52.26 & 50.50 & 48.95 & 57.11 & 51.05 & 43.83 & \textbf{63.64} & 78.68 & 72.10 & 78.29 & 75.92 & 71.97 & 80.92 & 80.39 & 79.60 & \textbf{85.26} \\
		\hline
		($\uparrow$) Ave.\footnotemark[1] & 43.19 & 48.20 & 51.42 & 50.79 & 48.86 & 58.20 & 52.25 & 44.35 & \textbf{63.75} & 78.57 & 71.23 & 78.81 & 76.07 & 70.78 & 78.90 & 81.10 & 79.46 & \textbf{84.97} \\
		\hline
	\end{tabular}
	\begin{tablenotes}
	    \item[1]($\uparrow$) Ave. denotes the average value over the above 10 domain sequences (i.e., S1, ..., S10).
    \end{tablenotes}
\end{threeparttable}}
\vspace{-0.2cm}
\caption{Previous task Evaluation: Average SC performance over 19 previous domains for each sequence.}
\label{tab:previousTask}
\vspace{-0.2cm}
\end{table*}

\textbf{Baselines:} We compare our LNB with NB, SVM~\cite{chang2011libsvm}, LSC~\cite{chen2015lifelong}, and Lifelong Voting (LLV)~\cite{xia2017distantly}. For LLV, we use its first voting method that can improve a past model using future knowledge. As traditional NB and SVM only work on a single domain data, we use their variations from \cite{chen2015lifelong}. These variations are called NB-T, NB-S, NB-ST, SVM-T, SVM-S and SVM-ST respectively, see our supplementary material. 

\textbf{Settings:} For NB, $\lambda$ is set to 0.1 for Lidstone smoothing. For SVM, we use the default parameters settings~\footnote{See: {\scriptsize{\url{http://www.csie.ntu.edu.tw/~cjlin/libsvm/}}}}. For LSC and LLV, we use their original parameters settings. For LNB, we empirically set $\gamma\!=\!2$ and $\sigma\!=\!3$. We use 5-fold cross validation in evaluation. For the dataset in the natural class distribution, we recorded the F1-score of the negative class. For the dataset in the balanced class distribution, we recorded the Accuracy of the positive and negative classes.

\textbf{New Task Evaluation}: Following \cite{chen2015lifelong}, each domain in the 20 domains data is treated as the new (target) domain with the rest 19 domains as the past domains. Table \ref{tab:newTask} shows the average results over the 20 domains. From the results, we make the following observations:
\vspace{-0.1cm}
\begin{enumerate*}
    \item Our LNB achieves the best F1-scores and Accuracy on two sets of datasets respectively. The results show the superiority of our LNB.
    \item NB-S (and SVM-S) is inferior to NB-T (and SVM-T), both of which are inferior to NB-ST (and SVM-ST). This shows that simply combining the training data from all past domains and the new domain is slightly beneficial, but worse than our model.
    \item LLV performs poorly as its voting does not fit our setting. In \cite{xia2017distantly}, all tasks are from the same domain, but our tasks are from different domains.
    \item Our model is slight better than LSC on the dataset in the balanced class distribution, but markedly better than LSC on the dataset in the natural class distribution~\footnote{In Table \ref{tab:newTask}, the F1-scores (56.62) for LSC is not the same as that reported in \cite{chen2015lifelong} because we used 80\% reviews of each past domain for training while \cite{chen2015lifelong} used all reviews for training because we need to test on past domains, while \cite{chen2015lifelong} does not do that.}. Note that this is not our main results as our main goal is to go back to help previous/past domain models without retraining, which LSC cannot do because LSC needs the past domain data to optimize its objective function.
    \vspace{-0.1cm}
\end{enumerate*}

\textbf{Previous Task Evaluation}: We now evaluate how each previous domain performs after some new/future domains have been learned. Since LSC cannot use the future knowledge to improve past domain models without retraining, for each past domain, we use the classifier built when the past domain was the new domain at that time. For NB-S, NB-ST, SVM-S and SVM-ST, we also use the classifier built at that time. For NB-T and SVM-T, we use the classifier built on each previous domain. For LLV, we use it as it was done because LLV is a voting method, which can use future models to vote in any past domain classification. We give the results after all 20 domains have been learned. Since in this case the ordering of domains may affect the experiment results, we randomly created 10 domain sequences. Due to the space limit, we provide the created 10 domain sequences in the supplementary material. For each sequence, the test results of the previous 19 domains are averaged and the average value is shown in Table \ref{tab:previousTask}.

From Table \ref{tab:previousTask}, we clearly see that LNB again outperforms all baselines. Although in helping future domain learning on the dataset in the balanced class distribution, LNB is only slightly better than LSC (see Table \ref{tab:newTask}), the ability of LNB to improve past domain models using future knowledge clearly shows its superiority to LSC. 

\begin{figure}[!htb]
    \centering
    \includegraphics[width=0.235\textwidth]{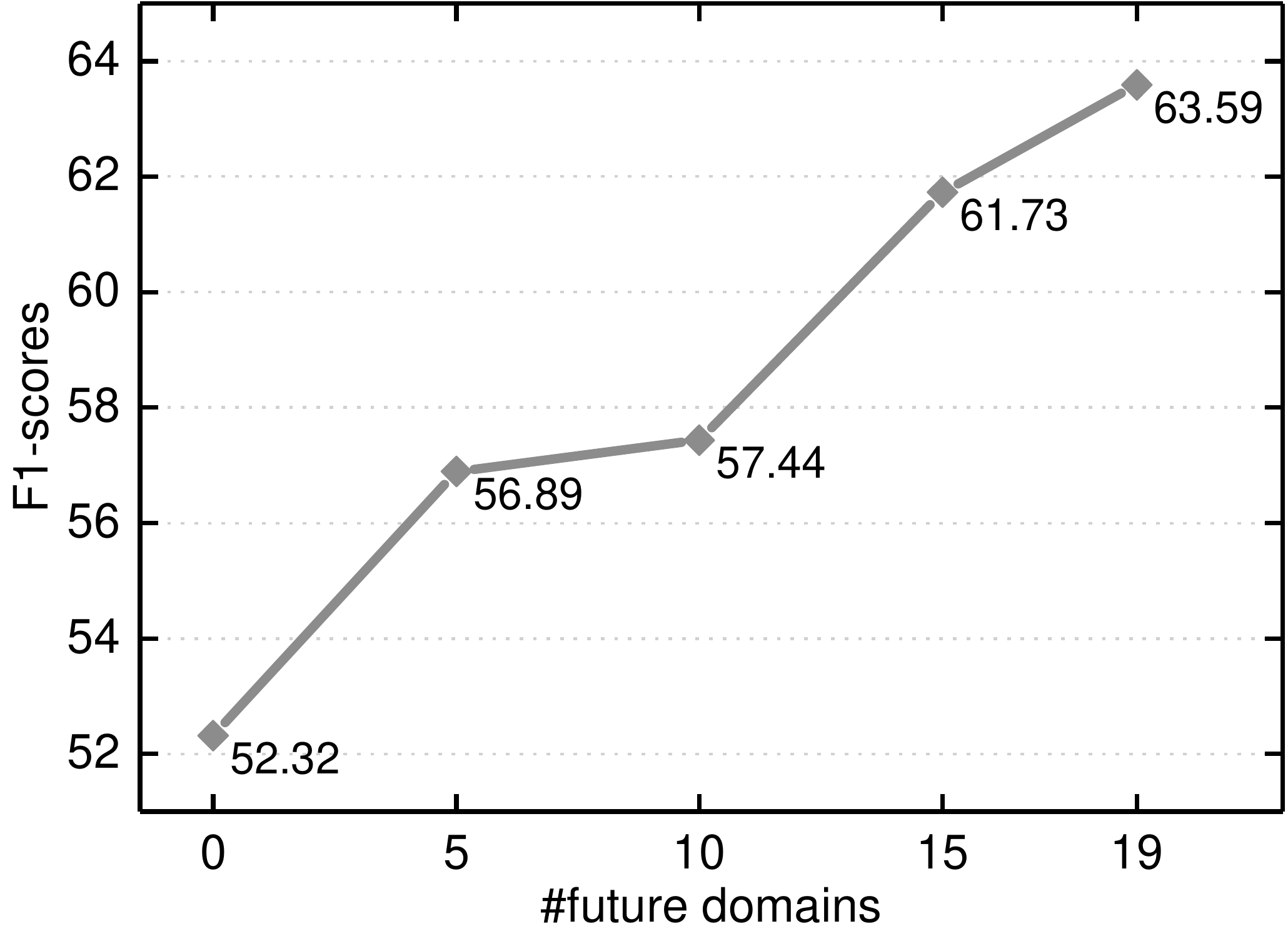}
    \hfill
    \includegraphics[width=0.235\textwidth]{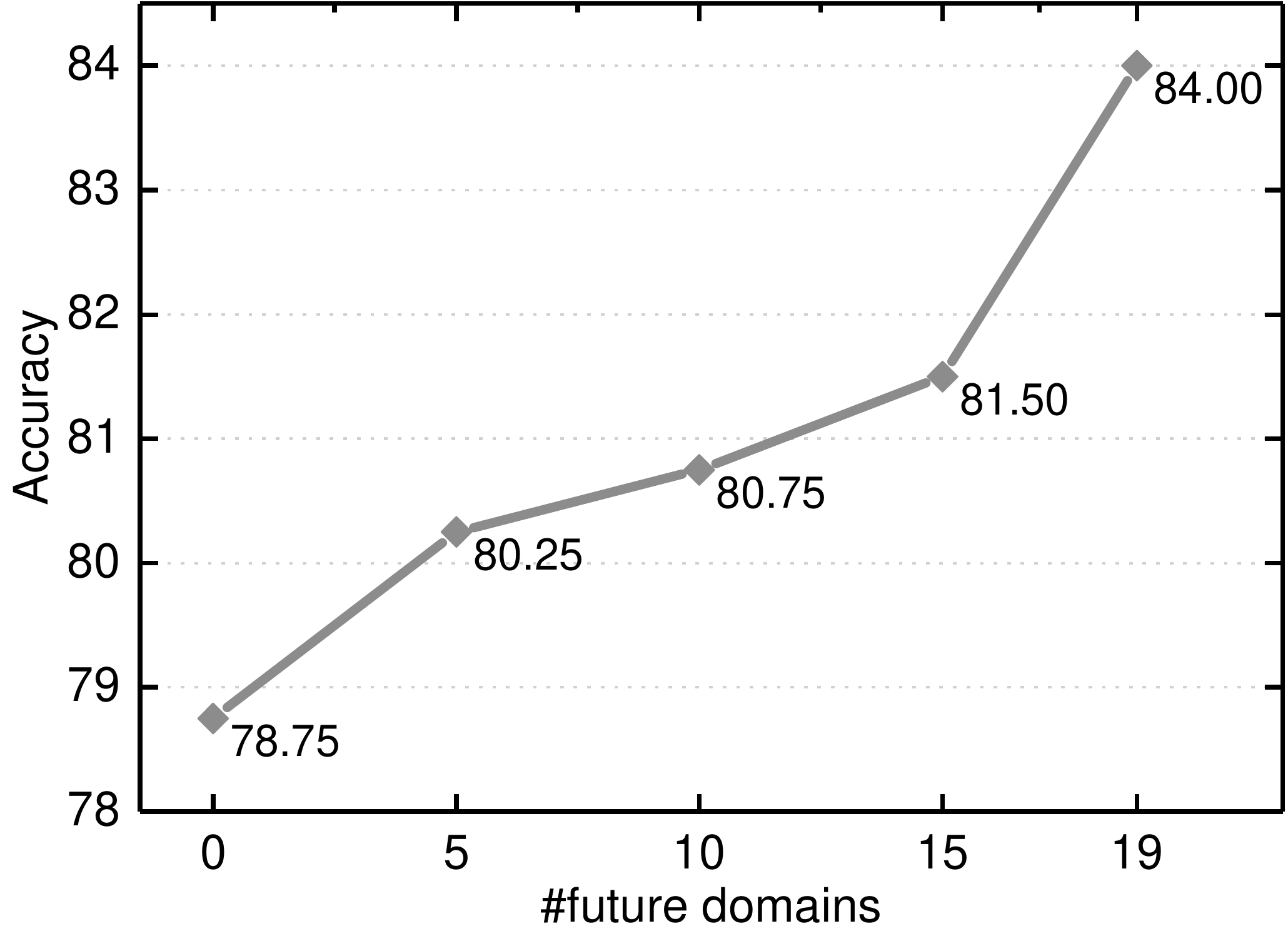}
    \vspace{-0.1cm}
    \caption{(Left): Effects of \#future domains on LNB in natural class distribution. (Right): Effects of \#future domains on LNB in balanced class distribution.}
    \label{fig:taskEffect}
\end{figure}

\textbf{Task Effect:} Here we further evaluate the performance of our LNB in helping past domain models using different number of new/future domains (denoted as \#future domains). In this experiment, we treat the first domain in each domain sequence as the target domain and vary the number of future domains. The curve of the average test results over 10 domain sequences is shown in Figure \ref{fig:taskEffect}. The curve clearly shows that LNB performs better with more future domains. This indicates that LNB indeed has the ability to go back to improve the past domain models using future knowledge.


\section{Conclusions}
This paper studied a new lifelong learning (LL) setting where the system uses the knowledge learned in future tasks to improve past task models with no retraining using the training data of the past tasks. We proposed a technique in this new LL setting by exploiting the generative model parameters of na\"ive Bayes. Experimental results showed the effectiveness of the approach. We believe this new setting is a promising direction for LL because we humans often learn new knowledge to solve past problems and future problems.

\appendix
\section{Appendices}
\label{sec:appendix}
Here we introduce the datasets, the baselines and the 10 domain sequences used in our experiments.

\textbf{Datasets:} It contains a collection of product reviews from 20 types of products or domains from Amazon.com. Each domain contains 1000 reviews. Each review has been assigned a sentiment label, i.e., positive (+) or negative (-), based on the rating score. The names of these 20 domains with a serial number for each domain and the proportion of negative reviews are shown in Table \ref{tab:domainName}.
\begin{table}[!htb]
\small
\renewcommand\arraystretch{1.3}
    \centering
    \resizebox{0.49\textwidth}{!}{
    \begin{tabular}{|ll|ll|ll|ll|}
    \hline
         $\raisebox{.5pt}{\textcircled{\raisebox{-.2pt}{\scriptsize{1}}}}$ Alarm Clock & 30.51 & $\raisebox{.5pt}{\textcircled{\raisebox{-.2pt}{\scriptsize{11}}}}$ Home Theater System & 28.84 \\ 
         $\raisebox{.5pt}{\textcircled{\raisebox{-.2pt}{\scriptsize{2}}}}$ Baby & 16.45 & $\raisebox{.5pt}{\textcircled{\raisebox{-.2pt}{\scriptsize{12}}}}$ Jewelry & 12.21 \\ 
         $\raisebox{.5pt}{\textcircled{\raisebox{-.2pt}{\scriptsize{3}}}}$ Bag & 11.97 & $\raisebox{.5pt}{\textcircled{\raisebox{-.2pt}{\scriptsize{13}}}}$ Keyboard & 22.66 \\ 
         $\raisebox{.5pt}{\textcircled{\raisebox{-.2pt}{\scriptsize{4}}}}$ Cable Modem & 12.53 & $\raisebox{.5pt}{\textcircled{\raisebox{-.2pt}{\scriptsize{14}}}}$ Magazine Subscriptions & 26.88 \\ 
         $\raisebox{.5pt}{\textcircled{\raisebox{-.2pt}{\scriptsize{5}}}}$ Dumbbell & 16.04 & $\raisebox{.5pt}{\textcircled{\raisebox{-.2pt}{\scriptsize{15}}}}$ Movies TV & 10.86 \\ 
         $\raisebox{.5pt}{\textcircled{\raisebox{-.2pt}{\scriptsize{6}}}}$ Flashlight & 11.69 & $\raisebox{.5pt}{\textcircled{\raisebox{-.2pt}{\scriptsize{16}}}}$ Projector & 20.24 \\ 
         $\raisebox{.5pt}{\textcircled{\raisebox{-.2pt}{\scriptsize{7}}}}$ Jewelry & 19.50 & $\raisebox{.5pt}{\textcircled{\raisebox{-.2pt}{\scriptsize{17}}}}$ Rice Cooker & 18.64 \\ 
         $\raisebox{.5pt}{\textcircled{\raisebox{-.2pt}{\scriptsize{8}}}}$ Gloves & 13.76 & $\raisebox{.5pt}{\textcircled{\raisebox{-.2pt}{\scriptsize{18}}}}$ Sandal & 12.11 \\ 
         $\raisebox{.5pt}{\textcircled{\raisebox{-.2pt}{\scriptsize{9}}}}$ Graphics Card & 14.58 & $\raisebox{.5pt}{\textcircled{\raisebox{-.2pt}{\scriptsize{19}}}}$ Vacuum & 22.07 \\ 
         $\raisebox{.5pt}{\textcircled{\raisebox{-.2pt}{\scriptsize{10}}}}$ Headphone & 20.99 & $\raisebox{.5pt}{\textcircled{\raisebox{-.2pt}{\scriptsize{20}}}}$ Video Games & 20.93 \\ 
    \hline
    \end{tabular}
    }
    \caption{Names of 20 domains with a serial number for each domain and the proportion of negative reviews in each domain.}
    \label{tab:domainName}
\end{table}

In our experiments, we use the following two sets of datasets with different class distribution:
\vspace{-0.1cm}
\begin{itemize*}
    \item \textbf{Natural class distribution}: We keep the natural distribution of the positive and negative reviews as shown in Table \ref{tab:domainName} and apply our system to the real-world situation. We use F-score of the negative class in evaluation as each domain has imbalanced class distribution and the number of reviews in the negative class is small. 
    \item \textbf{Balanced class distribution}: We also create a balanced dataset from the dataset in the natural class distribution. Each domain in the created dataset has 200 reviews (100 positive and 100 negative). This dataset is small because the number of negative reviews in each domain is small. We use Accuracy of both classes in evaluation as each domain has balanced class distribution. 
    \vspace{-0.1cm}
\end{itemize*}

\textbf{Baselines:} To have a comprehensive comparison, three variations of NB and SVM are created respectively:
\vspace{-0.1cm}
\begin{enumerate*}
	\item[1)] NB and SVM trained and tested on the target domain, denoted by \textbf{NB-T} and \textbf{SVM-T}.
	\item[2)] NB and SVM trained on the combined data from all non-target domains and tested on the target domain, denoted by \textbf{NB-S} and \textbf{SVM-S}.
	\item[3)] NB and SVM trained on the combined data from all domains (including the target domain) and tested on the target domain, denoted by \textbf{NB-ST} and \textbf{SVM-ST}.
	\vspace{-0.1cm}
\end{enumerate*}

We note that NB-T and SVM-T are traditional (non-lifelong learning) methods because they only use the data from the target task. NB-S, NB-ST, SVM-S and SVM-ST can be regarded as simple lifelong methods because they use the data from previous tasks.

\textbf{Domain Sequences:} Since the ordering of domains may affect the experiment results, we randomly created 10 domain sequences. The created 10 domain sequences are shown in Figure \ref{tab:sequence}.
\begin{figure}[!htb]
\renewcommand\arraystretch{1.1}
    \centering
    \resizebox{0.49\textwidth}{!}{
    \begin{tabular}{|r|}
    \hline
         S1: $\raisebox{.5pt}{\textcircled{\raisebox{-.2pt}{\scriptsize{17}}}}$$\raisebox{.5pt}{\textcircled{\raisebox{-.2pt}{\scriptsize{7}}}}$$\raisebox{.5pt}{\textcircled{\raisebox{-.2pt}{\scriptsize{16}}}}$$\raisebox{.5pt}{\textcircled{\raisebox{-.2pt}{\scriptsize{11}}}}$$\raisebox{.5pt}{\textcircled{\raisebox{-.2pt}{\scriptsize{10}}}}$$\raisebox{.5pt}{\textcircled{\raisebox{-.2pt}{\scriptsize{13}}}}$$\raisebox{.5pt}{\textcircled{\raisebox{-.2pt}{\scriptsize{12}}}}$$\raisebox{.5pt}{\textcircled{\raisebox{-.2pt}{\scriptsize{6}}}}$$\raisebox{.5pt}{\textcircled{\raisebox{-.2pt}{\scriptsize{5}}}}$$\raisebox{.5pt}{\textcircled{\raisebox{-.2pt}{\scriptsize{8}}}}$$\raisebox{.5pt}{\textcircled{\raisebox{-.2pt}{\scriptsize{18}}}}$$\raisebox{.5pt}{\textcircled{\raisebox{-.2pt}{\scriptsize{4}}}}$$\raisebox{.5pt}{\textcircled{\raisebox{-.2pt}{\scriptsize{14}}}}$$\raisebox{.5pt}{\textcircled{\raisebox{-.2pt}{\scriptsize{2}}}}$$\raisebox{.5pt}{\textcircled{\raisebox{-.2pt}{\scriptsize{9}}}}$$\raisebox{.5pt}{\textcircled{\raisebox{-.2pt}{\scriptsize{1}}}}$$\raisebox{.5pt}{\textcircled{\raisebox{-.2pt}{\scriptsize{15}}}}$$\raisebox{.5pt}{\textcircled{\raisebox{-.2pt}{\scriptsize{19}}}}$$\raisebox{.5pt}{\textcircled{\raisebox{-.2pt}{\scriptsize{3}}}}$$\raisebox{.5pt}{\textcircled{\raisebox{-.2pt}{\scriptsize{20}}}}$ \\
         S2: $\raisebox{.5pt}{\textcircled{\raisebox{-.2pt}{\scriptsize{14}}}}$$\raisebox{.5pt}{\textcircled{\raisebox{-.2pt}{\scriptsize{13}}}}$$\raisebox{.5pt}{\textcircled{\raisebox{-.2pt}{\scriptsize{19}}}}$$\raisebox{.5pt}{\textcircled{\raisebox{-.2pt}{\scriptsize{10}}}}$$\raisebox{.5pt}{\textcircled{\raisebox{-.2pt}{\scriptsize{2}}}}$$\raisebox{.5pt}{\textcircled{\raisebox{-.2pt}{\scriptsize{12}}}}$$\raisebox{.5pt}{\textcircled{\raisebox{-.2pt}{\scriptsize{20}}}}$$\raisebox{.5pt}{\textcircled{\raisebox{-.2pt}{\scriptsize{1}}}}$$\raisebox{.5pt}{\textcircled{\raisebox{-.2pt}{\scriptsize{16}}}}$$\raisebox{.5pt}{\textcircled{\raisebox{-.2pt}{\scriptsize{5}}}}$$\raisebox{.5pt}{\textcircled{\raisebox{-.2pt}{\scriptsize{11}}}}$$\raisebox{.5pt}{\textcircled{\raisebox{-.2pt}{\scriptsize{4}}}}$$\raisebox{.5pt}{\textcircled{\raisebox{-.2pt}{\scriptsize{8}}}}$$\raisebox{.5pt}{\textcircled{\raisebox{-.2pt}{\scriptsize{9}}}}$$\raisebox{.5pt}{\textcircled{\raisebox{-.2pt}{\scriptsize{2}}}}$$\raisebox{.5pt}{\textcircled{\raisebox{-.2pt}{\scriptsize{17}}}}$$\raisebox{.5pt}{\textcircled{\raisebox{-.2pt}{\scriptsize{7}}}}$$\raisebox{.5pt}{\textcircled{\raisebox{-.2pt}{\scriptsize{6}}}}$$\raisebox{.5pt}{\textcircled{\raisebox{-.2pt}{\scriptsize{15}}}}$$\raisebox{.5pt}{\textcircled{\raisebox{-.2pt}{\scriptsize{18}}}}$ \\
         S3: $\raisebox{.5pt}{\textcircled{\raisebox{-.2pt}{\scriptsize{5}}}}$$\raisebox{.5pt}{\textcircled{\raisebox{-.2pt}{\scriptsize{7}}}}$$\raisebox{.5pt}{\textcircled{\raisebox{-.2pt}{\scriptsize{17}}}}$$\raisebox{.5pt}{\textcircled{\raisebox{-.2pt}{\scriptsize{9}}}}$$\raisebox{.5pt}{\textcircled{\raisebox{-.2pt}{\scriptsize{4}}}}$$\raisebox{.5pt}{\textcircled{\raisebox{-.2pt}{\scriptsize{16}}}}$$\raisebox{.5pt}{\textcircled{\raisebox{-.2pt}{\scriptsize{15}}}}$$\raisebox{.5pt}{\textcircled{\raisebox{-.2pt}{\scriptsize{12}}}}$$\raisebox{.5pt}{\textcircled{\raisebox{-.2pt}{\scriptsize{10}}}}$$\raisebox{.5pt}{\textcircled{\raisebox{-.2pt}{\scriptsize{14}}}}$$\raisebox{.5pt}{\textcircled{\raisebox{-.2pt}{\scriptsize{11}}}}$$\raisebox{.5pt}{\textcircled{\raisebox{-.2pt}{\scriptsize{8}}}}$$\raisebox{.5pt}{\textcircled{\raisebox{-.2pt}{\scriptsize{19}}}}$$\raisebox{.5pt}{\textcircled{\raisebox{-.2pt}{\scriptsize{18}}}}$$\raisebox{.5pt}{\textcircled{\raisebox{-.2pt}{\scriptsize{6}}}}$$\raisebox{.5pt}{\textcircled{\raisebox{-.2pt}{\scriptsize{13}}}}$$\raisebox{.5pt}{\textcircled{\raisebox{-.2pt}{\scriptsize{2}}}}$$\raisebox{.5pt}{\textcircled{\raisebox{-.2pt}{\scriptsize{3}}}}$$\raisebox{.5pt}{\textcircled{\raisebox{-.2pt}{\scriptsize{20}}}}$$\raisebox{.5pt}{\textcircled{\raisebox{-.2pt}{\scriptsize{1}}}}$ \\
         S4: $\raisebox{.5pt}{\textcircled{\raisebox{-.2pt}{\scriptsize{14}}}}$$\raisebox{.5pt}{\textcircled{\raisebox{-.2pt}{\scriptsize{4}}}}$$\raisebox{.5pt}{\textcircled{\raisebox{-.2pt}{\scriptsize{20}}}}$$\raisebox{.5pt}{\textcircled{\raisebox{-.2pt}{\scriptsize{7}}}}$$\raisebox{.5pt}{\textcircled{\raisebox{-.2pt}{\scriptsize{17}}}}$$\raisebox{.5pt}{\textcircled{\raisebox{-.2pt}{\scriptsize{15}}}}$$\raisebox{.5pt}{\textcircled{\raisebox{-.2pt}{\scriptsize{16}}}}$$\raisebox{.5pt}{\textcircled{\raisebox{-.2pt}{\scriptsize{8}}}}$$\raisebox{.5pt}{\textcircled{\raisebox{-.2pt}{\scriptsize{13}}}}$$\raisebox{.5pt}{\textcircled{\raisebox{-.2pt}{\scriptsize{9}}}}$$\raisebox{.5pt}{\textcircled{\raisebox{-.2pt}{\scriptsize{12}}}}$$\raisebox{.5pt}{\textcircled{\raisebox{-.2pt}{\scriptsize{6}}}}$$\raisebox{.5pt}{\textcircled{\raisebox{-.2pt}{\scriptsize{2}}}}$$\raisebox{.5pt}{\textcircled{\raisebox{-.2pt}{\scriptsize{3}}}}$$\raisebox{.5pt}{\textcircled{\raisebox{-.2pt}{\scriptsize{5}}}}$$\raisebox{.5pt}{\textcircled{\raisebox{-.2pt}{\scriptsize{10}}}}$$\raisebox{.5pt}{\textcircled{\raisebox{-.2pt}{\scriptsize{18}}}}$$\raisebox{.5pt}{\textcircled{\raisebox{-.2pt}{\scriptsize{11}}}}$$\raisebox{.5pt}{\textcircled{\raisebox{-.2pt}{\scriptsize{1}}}}$$\raisebox{.5pt}{\textcircled{\raisebox{-.2pt}{\scriptsize{19}}}}$ \\
         S5: $\raisebox{.5pt}{\textcircled{\raisebox{-.2pt}{\scriptsize{5}}}}$$\raisebox{.5pt}{\textcircled{\raisebox{-.2pt}{\scriptsize{16}}}}$$\raisebox{.5pt}{\textcircled{\raisebox{-.2pt}{\scriptsize{2}}}}$$\raisebox{.5pt}{\textcircled{\raisebox{-.2pt}{\scriptsize{20}}}}$$\raisebox{.5pt}{\textcircled{\raisebox{-.2pt}{\scriptsize{18}}}}$$\raisebox{.5pt}{\textcircled{\raisebox{-.2pt}{\scriptsize{8}}}}$$\raisebox{.5pt}{\textcircled{\raisebox{-.2pt}{\scriptsize{13}}}}$$\raisebox{.5pt}{\textcircled{\raisebox{-.2pt}{\scriptsize{4}}}}$$\raisebox{.5pt}{\textcircled{\raisebox{-.2pt}{\scriptsize{6}}}}$$\raisebox{.5pt}{\textcircled{\raisebox{-.2pt}{\scriptsize{9}}}}$$\raisebox{.5pt}{\textcircled{\raisebox{-.2pt}{\scriptsize{10}}}}$$\raisebox{.5pt}{\textcircled{\raisebox{-.2pt}{\scriptsize{19}}}}$$\raisebox{.5pt}{\textcircled{\raisebox{-.2pt}{\scriptsize{7}}}}$$\raisebox{.5pt}{\textcircled{\raisebox{-.2pt}{\scriptsize{3}}}}$$\raisebox{.5pt}{\textcircled{\raisebox{-.2pt}{\scriptsize{11}}}}$$\raisebox{.5pt}{\textcircled{\raisebox{-.2pt}{\scriptsize{1}}}}$$\raisebox{.5pt}{\textcircled{\raisebox{-.2pt}{\scriptsize{14}}}}$$\raisebox{.5pt}{\textcircled{\raisebox{-.2pt}{\scriptsize{15}}}}$$\raisebox{.5pt}{\textcircled{\raisebox{-.2pt}{\scriptsize{12}}}}$$\raisebox{.5pt}{\textcircled{\raisebox{-.2pt}{\scriptsize{17}}}}$ \\
         S6: $\raisebox{.5pt}{\textcircled{\raisebox{-.2pt}{\scriptsize{15}}}}$$\raisebox{.5pt}{\textcircled{\raisebox{-.2pt}{\scriptsize{11}}}}$$\raisebox{.5pt}{\textcircled{\raisebox{-.2pt}{\scriptsize{4}}}}$$\raisebox{.5pt}{\textcircled{\raisebox{-.2pt}{\scriptsize{20}}}}$$\raisebox{.5pt}{\textcircled{\raisebox{-.2pt}{\scriptsize{17}}}}$$\raisebox{.5pt}{\textcircled{\raisebox{-.2pt}{\scriptsize{3}}}}$$\raisebox{.5pt}{\textcircled{\raisebox{-.2pt}{\scriptsize{7}}}}$$\raisebox{.5pt}{\textcircled{\raisebox{-.2pt}{\scriptsize{10}}}}$$\raisebox{.5pt}{\textcircled{\raisebox{-.2pt}{\scriptsize{16}}}}$$\raisebox{.5pt}{\textcircled{\raisebox{-.2pt}{\scriptsize{12}}}}$$\raisebox{.5pt}{\textcircled{\raisebox{-.2pt}{\scriptsize{18}}}}$$\raisebox{.5pt}{\textcircled{\raisebox{-.2pt}{\scriptsize{1}}}}$$\raisebox{.5pt}{\textcircled{\raisebox{-.2pt}{\scriptsize{2}}}}$$\raisebox{.5pt}{\textcircled{\raisebox{-.2pt}{\scriptsize{13}}}}$$\raisebox{.5pt}{\textcircled{\raisebox{-.2pt}{\scriptsize{5}}}}$$\raisebox{.5pt}{\textcircled{\raisebox{-.2pt}{\scriptsize{8}}}}$$\raisebox{.5pt}{\textcircled{\raisebox{-.2pt}{\scriptsize{19}}}}$$\raisebox{.5pt}{\textcircled{\raisebox{-.2pt}{\scriptsize{6}}}}$$\raisebox{.5pt}{\textcircled{\raisebox{-.2pt}{\scriptsize{9}}}}$$\raisebox{.5pt}{\textcircled{\raisebox{-.2pt}{\scriptsize{14}}}}$ \\
         S7: $\raisebox{.5pt}{\textcircled{\raisebox{-.2pt}{\scriptsize{20}}}}$$\raisebox{.5pt}{\textcircled{\raisebox{-.2pt}{\scriptsize{13}}}}$$\raisebox{.5pt}{\textcircled{\raisebox{-.2pt}{\scriptsize{2}}}}$$\raisebox{.5pt}{\textcircled{\raisebox{-.2pt}{\scriptsize{15}}}}$$\raisebox{.5pt}{\textcircled{\raisebox{-.2pt}{\scriptsize{9}}}}$$\raisebox{.5pt}{\textcircled{\raisebox{-.2pt}{\scriptsize{17}}}}$$\raisebox{.5pt}{\textcircled{\raisebox{-.2pt}{\scriptsize{14}}}}$$\raisebox{.5pt}{\textcircled{\raisebox{-.2pt}{\scriptsize{5}}}}$$\raisebox{.5pt}{\textcircled{\raisebox{-.2pt}{\scriptsize{16}}}}$$\raisebox{.5pt}{\textcircled{\raisebox{-.2pt}{\scriptsize{18}}}}$$\raisebox{.5pt}{\textcircled{\raisebox{-.2pt}{\scriptsize{7}}}}$$\raisebox{.5pt}{\textcircled{\raisebox{-.2pt}{\scriptsize{4}}}}$$\raisebox{.5pt}{\textcircled{\raisebox{-.2pt}{\scriptsize{11}}}}$$\raisebox{.5pt}{\textcircled{\raisebox{-.2pt}{\scriptsize{6}}}}$$\raisebox{.5pt}{\textcircled{\raisebox{-.2pt}{\scriptsize{1}}}}$$\raisebox{.5pt}{\textcircled{\raisebox{-.2pt}{\scriptsize{8}}}}$$\raisebox{.5pt}{\textcircled{\raisebox{-.2pt}{\scriptsize{10}}}}$$\raisebox{.5pt}{\textcircled{\raisebox{-.2pt}{\scriptsize{3}}}}$$\raisebox{.5pt}{\textcircled{\raisebox{-.2pt}{\scriptsize{19}}}}$$\raisebox{.5pt}{\textcircled{\raisebox{-.2pt}{\scriptsize{12}}}}$ \\
         S8: $\raisebox{.5pt}{\textcircled{\raisebox{-.2pt}{\scriptsize{18}}}}$$\raisebox{.5pt}{\textcircled{\raisebox{-.2pt}{\scriptsize{11}}}}$$\raisebox{.5pt}{\textcircled{\raisebox{-.2pt}{\scriptsize{6}}}}$$\raisebox{.5pt}{\textcircled{\raisebox{-.2pt}{\scriptsize{12}}}}$$\raisebox{.5pt}{\textcircled{\raisebox{-.2pt}{\scriptsize{13}}}}$$\raisebox{.5pt}{\textcircled{\raisebox{-.2pt}{\scriptsize{1}}}}$$\raisebox{.5pt}{\textcircled{\raisebox{-.2pt}{\scriptsize{5}}}}$$\raisebox{.5pt}{\textcircled{\raisebox{-.2pt}{\scriptsize{7}}}}$$\raisebox{.5pt}{\textcircled{\raisebox{-.2pt}{\scriptsize{3}}}}$$\raisebox{.5pt}{\textcircled{\raisebox{-.2pt}{\scriptsize{16}}}}$$\raisebox{.5pt}{\textcircled{\raisebox{-.2pt}{\scriptsize{2}}}}$$\raisebox{.5pt}{\textcircled{\raisebox{-.2pt}{\scriptsize{4}}}}$$\raisebox{.5pt}{\textcircled{\raisebox{-.2pt}{\scriptsize{14}}}}$$\raisebox{.5pt}{\textcircled{\raisebox{-.2pt}{\scriptsize{8}}}}$$\raisebox{.5pt}{\textcircled{\raisebox{-.2pt}{\scriptsize{15}}}}$$\raisebox{.5pt}{\textcircled{\raisebox{-.2pt}{\scriptsize{19}}}}$$\raisebox{.5pt}{\textcircled{\raisebox{-.2pt}{\scriptsize{17}}}}$$\raisebox{.5pt}{\textcircled{\raisebox{-.2pt}{\scriptsize{10}}}}$$\raisebox{.5pt}{\textcircled{\raisebox{-.2pt}{\scriptsize{9}}}}$$\raisebox{.5pt}{\textcircled{\raisebox{-.2pt}{\scriptsize{20}}}}$ \\
         S9: $\raisebox{.5pt}{\textcircled{\raisebox{-.2pt}{\scriptsize{16}}}}$$\raisebox{.5pt}{\textcircled{\raisebox{-.2pt}{\scriptsize{2}}}}$$\raisebox{.5pt}{\textcircled{\raisebox{-.2pt}{\scriptsize{10}}}}$$\raisebox{.5pt}{\textcircled{\raisebox{-.2pt}{\scriptsize{9}}}}$$\raisebox{.5pt}{\textcircled{\raisebox{-.2pt}{\scriptsize{12}}}}$$\raisebox{.5pt}{\textcircled{\raisebox{-.2pt}{\scriptsize{19}}}}$$\raisebox{.5pt}{\textcircled{\raisebox{-.2pt}{\scriptsize{6}}}}$$\raisebox{.5pt}{\textcircled{\raisebox{-.2pt}{\scriptsize{11}}}}$$\raisebox{.5pt}{\textcircled{\raisebox{-.2pt}{\scriptsize{18}}}}$$\raisebox{.5pt}{\textcircled{\raisebox{-.2pt}{\scriptsize{20}}}}$$\raisebox{.5pt}{\textcircled{\raisebox{-.2pt}{\scriptsize{4}}}}$$\raisebox{.5pt}{\textcircled{\raisebox{-.2pt}{\scriptsize{1}}}}$$\raisebox{.5pt}{\textcircled{\raisebox{-.2pt}{\scriptsize{3}}}}$$\raisebox{.5pt}{\textcircled{\raisebox{-.2pt}{\scriptsize{17}}}}$$\raisebox{.5pt}{\textcircled{\raisebox{-.2pt}{\scriptsize{5}}}}$$\raisebox{.5pt}{\textcircled{\raisebox{-.2pt}{\scriptsize{13}}}}$$\raisebox{.5pt}{\textcircled{\raisebox{-.2pt}{\scriptsize{15}}}}$$\raisebox{.5pt}{\textcircled{\raisebox{-.2pt}{\scriptsize{8}}}}$$\raisebox{.5pt}{\textcircled{\raisebox{-.2pt}{\scriptsize{14}}}}$$\raisebox{.5pt}{\textcircled{\raisebox{-.2pt}{\scriptsize{7}}}}$ \\
         S10: $\raisebox{.5pt}{\textcircled{\raisebox{-.2pt}{\scriptsize{12}}}}$$\raisebox{.5pt}{\textcircled{\raisebox{-.2pt}{\scriptsize{16}}}}$$\raisebox{.5pt}{\textcircled{\raisebox{-.2pt}{\scriptsize{6}}}}$$\raisebox{.5pt}{\textcircled{\raisebox{-.2pt}{\scriptsize{8}}}}$$\raisebox{.5pt}{\textcircled{\raisebox{-.2pt}{\scriptsize{19}}}}$$\raisebox{.5pt}{\textcircled{\raisebox{-.2pt}{\scriptsize{2}}}}$$\raisebox{.5pt}{\textcircled{\raisebox{-.2pt}{\scriptsize{7}}}}$$\raisebox{.5pt}{\textcircled{\raisebox{-.2pt}{\scriptsize{1}}}}$$\raisebox{.5pt}{\textcircled{\raisebox{-.2pt}{\scriptsize{13}}}}$$\raisebox{.5pt}{\textcircled{\raisebox{-.2pt}{\scriptsize{17}}}}$$\raisebox{.5pt}{\textcircled{\raisebox{-.2pt}{\scriptsize{3}}}}$$\raisebox{.5pt}{\textcircled{\raisebox{-.2pt}{\scriptsize{9}}}}$$\raisebox{.5pt}{\textcircled{\raisebox{-.2pt}{\scriptsize{4}}}}$$\raisebox{.5pt}{\textcircled{\raisebox{-.2pt}{\scriptsize{11}}}}$$\raisebox{.5pt}{\textcircled{\raisebox{-.2pt}{\scriptsize{18}}}}$$\raisebox{.5pt}{\textcircled{\raisebox{-.2pt}{\scriptsize{15}}}}$$\raisebox{.5pt}{\textcircled{\raisebox{-.2pt}{\scriptsize{20}}}}$$\raisebox{.5pt}{\textcircled{\raisebox{-.2pt}{\scriptsize{5}}}}$$\raisebox{.5pt}{\textcircled{\raisebox{-.2pt}{\scriptsize{10}}}}$$\raisebox{.5pt}{\textcircled{\raisebox{-.2pt}{\scriptsize{14}}}}$ \\
    \hline
    \end{tabular}
    }
    \caption{The randomly created 10 domain sequences.}
    \label{tab:sequence}
    \vspace{-0.2cm}
\end{figure}

\bibliographystyle{acl_natbib}

\end{document}